\documentclass{article}
\usepackage{spconf,amsmath,graphicx}
\usepackage[ruled,vlined]{algorithm2e}

\usepackage{enumitem}
\setlist{nosep, leftmargin=14pt}

\usepackage{mwe} 

\title{DerMAE: Improving skin lesion classification through conditioned latent diffusion and MAE distillation}
\name{
\parbox{\linewidth}{\centering
Francisco Filho \qquad Kelvin Cunha \qquad Fábio Papais \qquad Emanoel dos Santos \qquad Rodrigo Mota \qquad Thales Bezerra \qquad Erico Medeiros \qquad Paulo Borba \qquad Tsang Ing Ren}}

\address{Centro de Informática, Universidade Federal de Pernambuco, Brazil}
     
\begin{document}
%
\maketitle
\begin{abstract}
Skin lesion classification datasets often suffer from severe class imbalance, with malignant cases significantly underrepresented, leading to biased decision boundaries during deep learning training. We address this challenge using class-conditioned diffusion models to generate synthetic dermatological images, followed by self-supervised MAE pretraining to enable huge ViT models to learn robust, domain-relevant features. To support deployment in practical clinical settings, where lightweight models are required, we apply knowledge distillation to transfer these representations to a smaller ViT student suitable for mobile devices. Our results show that MAE pretraining on synthetic data, combined with distillation, improves classification performance while enabling efficient on-device inference for practical clinical use.

\end{abstract}

\section{Introduction}
\label{sec:intro}

Skin cancer is among the most prevalent neoplasias worldwide, yet few patients actively seek or have access to specialized medical attention~\cite{tsang2006even}. Initial assessments are often performed by general practitioners, who may struggle to reach accurate diagnoses without specialist support, while biopsies are invasive and typically avoided for suspected benign lesions. Moreover, the lack of centralized and standardized patient data, together with variability related to physical traits, geography, or ethnicity, contributes to biased dermatology datasets. As a result, most medical imaging datasets are small and highly imbalanced, leading deep learning models to favor majority classes and systematically misclassify underrepresented lesions~\cite{ding2025111680}. This issue is particularly pronounced in dermatology, where benign lesions dominate. To alleviate this limitation, data augmentation techniques have been widely adopted~\cite{patil2025enhancing}. Conventional methods rely on geometric and color transformations~\cite{perez2018data}, but they are restricted to modifying existing samples and cannot introduce novel lesion patterns. Synthetic image generation has therefore emerged as a promising alternative~\cite{farooq2024derm, patil2025enhancing}, with recent diffusion models~\cite{ho2020denoising} showing potential to mitigate class imbalance through data synthesis~\cite{ding2025111680}. Meanwhile, in classification, architectures such as the Vision Transformer (ViT)~\cite{touvron2021training} have demonstrated powerful representation learning capabilities via self-attention; however, their strong dependence on large-scale data significantly limits their effectiveness on current dermatology datasets.

In this work, we address class imbalance in skin lesion classification using synthetic images generated by class-conditioned diffusion models. We introduce a pre-training strategy based on Masked Autoencoders (MAE), where the model is first trained on large-scale synthetic data to learn generalizable representations and supplement underrepresented malignant classes. To support deployment in resource-constrained clinical settings, we further apply knowledge distillation, using the MAE as a teacher to train smaller classifiers suitable for mobile devices. Experimental results demonstrate that synthetic pre-training combined with distillation improves classification performance while enabling compact, point-of-care dermatology models.

\section{Methodology}
\label{sec:method}

\subsection{Synthetic Data Generation}

\begin{figure*}
	\centering
	\includegraphics[width=1.0 \linewidth]{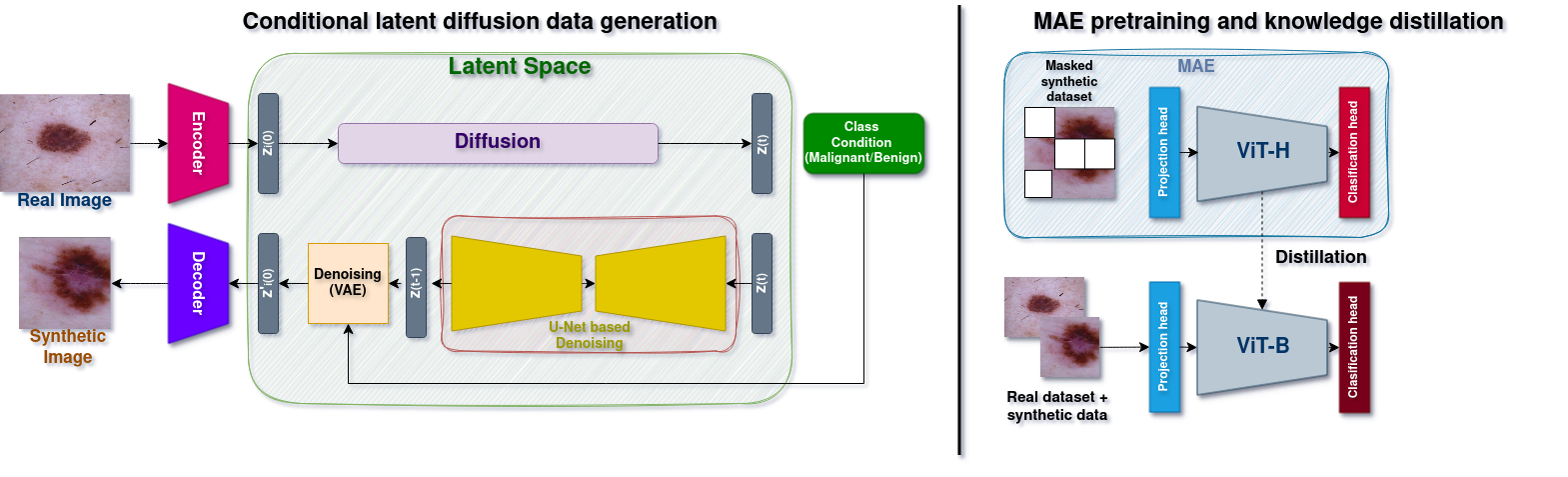}
	\caption{Overview of the proposed framework. First, class-conditioned latent diffusion generates synthetic skin image. Next, the synthetic data is used to pretrain a ViT-H model using an MAE objective. Finally, knowledge distillation tunes a smaller student model from the pretrained ViT, which is fine-tuned using a combination of real and synthetic data.}
	\label{fig:overview}
\end{figure*}

We evaluate our approach on the HAM10000 dermoscopic dataset~\cite{tschandl2018ham10000}, which contains $10{,}000$ images from eight lesion categories, including three malignant classes. The dataset is highly imbalanced, with approximately $90\%$ benign samples. To mitigate this imbalance, we augment the training data with synthetic images generated using a Denoising Diffusion Probabilistic Model (DDPM)~\cite{ho2020denoising}. Two generation strategies are considered: (i) \textbf{unconditional generation}, in which $600{,}000$ samples are generated without class conditioning to model the overall lesion distribution; and (ii) \textbf{conditional generation}, where class embeddings are introduced during inference to explicitly balance malignant and benign classes.

The model employs a U-Net backbone~\cite{ho2020denoising} to parameterize the denoising function $\epsilon_{\theta}(\mathbf{x}_t, t)$, following the standard DDPM formulation~\cite{ho2020denoising}. The forward pass progressively corrupts a clean image $\mathbf{x}_0$ into a noisy latent variable $\mathbf{x}_t$ through the addition of Gaussian noise (Equation~\ref{eq:noising}). Where $\{\alpha_t\}_{t=1}^{T}$ follows a linear noise schedule with $T = 1000$ timesteps. Initially, the model was trained to predict the noise component via the mean squared error objective, as in Equation~\ref{eq:mse}.

\begin{equation}
\label{eq:noising}
q(\mathbf{x}_t \mid \mathbf{x}_0) = \mathcal{N}\left(\mathbf{x}_t; \sqrt{\alpha_t}\,\mathbf{x}_0,\,(1 - \alpha_t)\mathbf{I}\right)
\end{equation}

\begin{equation}
\label{eq:mse}
\mathcal{L}_{MSE}(\theta) = \mathbf{E}_{\mathbf{x}_0, t, \epsilon \sim \mathcal{N}(0,\mathbf{I})}\left[\left\|\,\epsilon - \epsilon_{\theta}(\mathbf{x}_t, t)\,\right\|_2^2\right]
\end{equation}

To improve perceptual fidelity and reduce low-frequency artifacts commonly introduced by the MSE objective, we incorporated a perceptual loss term~\cite{johnson2016perceptual}. The final training objective is illustrated in Equation~\ref{eq:total_loss}.

\begin{equation}
\label{eq:total_loss}
\mathcal{L}(\theta) = \mathcal{L}_{\text{MSE}}(\theta) + \lambda\,\mathcal{L}_{\text{perc}}(\theta)
\end{equation}

Training was performed using the Adam optimizer (learning rate $1\times10^{-4}$), and resolution of $256 \times 256$ pixels. Sampling followed the standard stochastic DDPM denoising to retrieve synthetic samples from the learned distribution.

\subsection{Self-supervised pre-training}

To leverage both the original and synthetically generated data in a self-supervised manner, we employ a Masked Autoencoder (MAE)~\cite{mae} with a ViT-Huge backbone (ViT-H/16). The model employs a patch size of \(16 \times 16\), an embedding dimension of $1024$, $24$ transformer encoder layers, and $16$ self-attention heads. During training, a random masking ratio of $75\%$ is applied to the input patches, such that only a subset of visible tokens is processed by the encoder, while the decoder is tasked with reconstructing the missing patches. This reconstruction objective encourages the model to learn semantically meaningful latent representations that capture global lesion structures rather than low-level correlations. Formally, let \( \mathbf{x} \in \mathbf{R}^{H \times W \times 3} \) denote an input image divided into \(N\) patches. A binary mask \(M \in \{0,1\}^N\) selects the subset of visible patches for the encoder, yielding encoded tokens \(z = f_{\theta}(\mathbf{x} \odot M)\), where \(f_\theta\) is the ViT encoder. The decoder \(g_{\phi}\) then reconstructs the original image patches \(\hat{\mathbf{x}}\) from \(z\) and mask tokens. The training objective is a pixel-wise reconstruction loss over the masked regions, as shown in Equation~\ref{eq:objective_vit}.

\begin{equation}
\label{eq:objective_vit}
\mathcal{L}_{\text{MAE}}(\theta, \phi) = 
\left\| \left(\mathbf{x} - g_{\phi}(f_{\theta}(\mathbf{x} \odot M))\right) \odot (1 - M) \right\|_2^2
\end{equation}

Training is performed for 100 epochs using the AdamW optimizer with cosine learning rate decay, using an initial learning rate of \(1 \times 10^{-4}\), batch size of 256, and weight decay of 0.05. The MAE pretraining stage enables the learning of generalized visual features from the large synthetic data, alleviating the limitations imposed by imbalance. These learned representations are subsequently transferred to smaller downstream classification models through knowledge distillation, allowing compact models to benefit from the richer feature space acquired during MAE pretraining.

\subsection{Knowledge distillation}

To take advantage of the representational capabilities acquired during MAE pretraining, we combine self-supervised learning with knowledge distillation to transfer knowledge from the ViT-H model to a smaller student network. Specifically, the encoder of the pretrained MAE (ViT-H) is adopted, and its output representations are used to guide a ViT-Base (ViT-B/16) student model. The student architecture consists of 12 transformer layers, an embedding dimension of 768, and 12 self-attention heads. Knowledge transfer is performed using a soft-target distillation strategy, wherein the student is encouraged to match both the teacher’s output distribution and the ground-truth class labels. Let \( p_{t} \) denote the teacher’s predicted probability distribution and \( p_{s} \) the student’s output, the distillation loss ($\mathcal{L}_{\text{KD}}$) is defined as in equation~\ref{eq:distill-loss}. \( \mathcal{L}_{\text{CE}} \) is the cross-entropy loss with ground-truth labels \(y\), \( \mathcal{L}_{\text{KL}} \) is the Kullback–Leibler divergence between softened probability distributions, and \( \alpha \) balances the contribution of soft and hard supervision. This distillation procedure enables the student network to inherit the high-level semantic structure and feature representations learned during MAE self-supervised pretraining. Our architecture is illustrated in Figure~\ref{fig:overview}.

\begin{equation}
\label{eq:distill-loss}
\mathcal{L}_{\text{KD}} = (1 - \alpha)\, \mathcal{L}_{\text{CE}}(p_{s}, y) \;+\; \alpha\, \mathcal{L}_{\text{KL}}(p_{s}, p_{t})
\end{equation}

\subsection{Experiments details}

Experiments were performed under three configurations to evaluate the effect of synthetic augmentation: (i) a \textbf{baseline} trained solely on HAM10000; (ii) \textbf{MAE + synthetic}, using MAE pre-training followed by training with synthetic samples to balance benign and malignant classes; and (iii) \textbf{distillation + augmentation (distill)}, combining real data with synthetic samples distilled from MAE pre-training.

The ViT-B/16 baseline was trained for 1000 epochs using AdamW (learning rate \(3 \times 10^{-5}\), batch size $64$, and weight decay $0.01$) with a cosine scheduler. Results were evaluated using accuracy and F1-score~\cite{tschandl2018ham10000, farooq2024derm} and compared against EfficientNet variants~\cite{tan2019efficientnet} and Derm-t2im~\cite{farooq2024derm}, which employs a Stable Diffusion-based framework for synthetic generation and subsequent binary classification.

\section{Experiments and discussions}

We trained EfficientNet and ViT baseline models using the original HAM10000 training split. From the $10,015$ available samples, we allocate $80\%$ for training ($8,012$) and $20\%$ for validation ($2,003$), retaining the original test set ($1,512$ images) for evaluation. It is important to note that the same training set is used for optimizing synthetic image generation. Consequently, no contamination occurs from including validation or test samples in the synthesis process. 

As shown in Table~\ref{tab:ham10000}, the ViT-based models exhibit limited performance, which is expected given that vision transformers require substantially larger datasets to learn stable representations. The limited size and severe class imbalance negatively affect their ability to generalize. In contrast, EfficientNet performs comparatively better. For all models, performance is higher in binary (malignant/benign) classification, reflecting the greater difficulty posed by the imbalance in multi-class settings. 

Notably, incorporating MAE pretraining on synthetic data consistently improves the performance of ViT architectures. In particular, F1-based metrics benefit the most, as the inclusion of synthetic samples helps balance the dataset, enabling models to learn malignant features more effectively and distinguish between lesion types. Furthermore, when applying knowledge distillation (distill) from the MAE-pretrained teacher, and fine-tuning with a combination of real and class-conditioned synthetic samples (synth) generated by our diffusion model, both ViT-B and EfficientNet-B0 achieve additional gains. These results demonstrate the effectiveness of synthetic-data-driven pretraining and distillation in mitigating data scarcity and imbalance in dermatological image classification. 

\begin{figure}
	\centering
	\includegraphics[width=1.0 \linewidth]{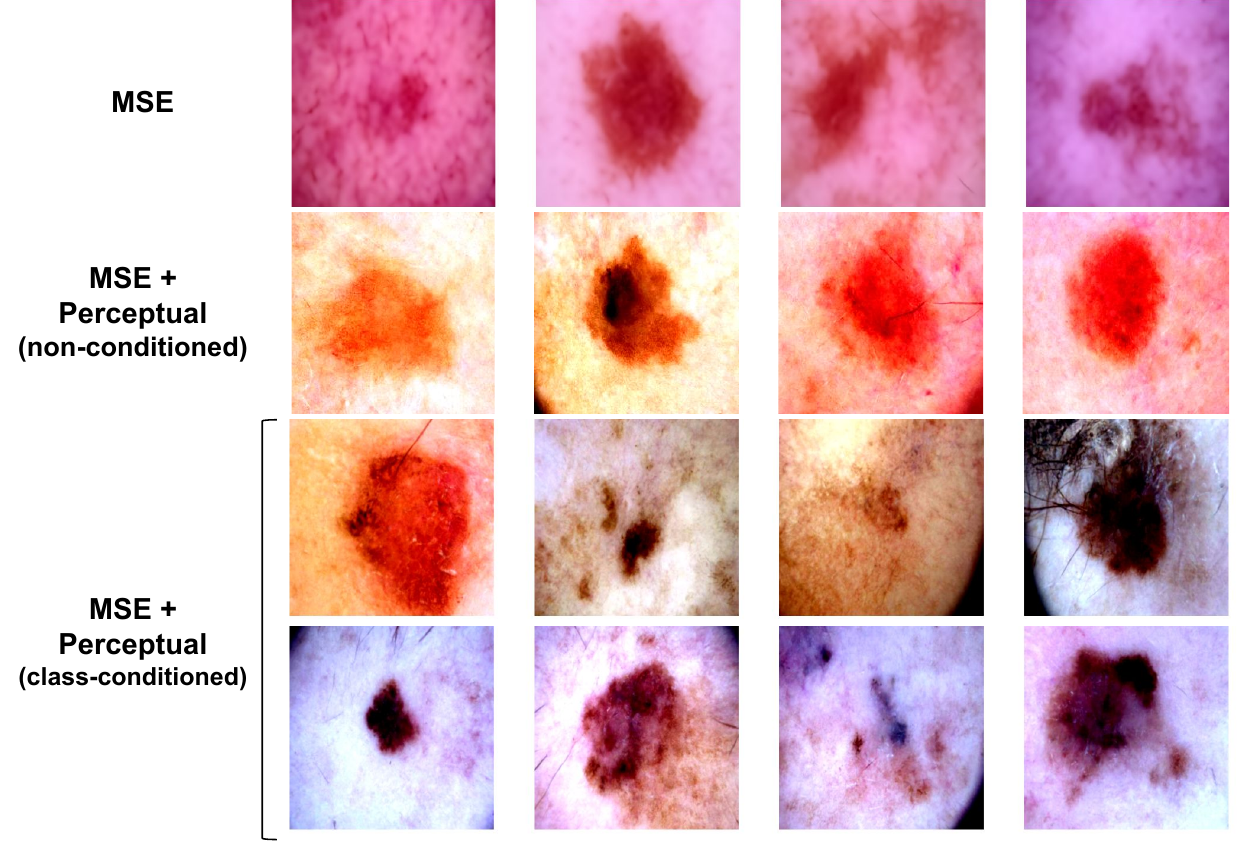}
	\caption{Qualitative comparison of latent diffusion generation strategies. Rows (top to bottom): MSE loss, MSE + perceptual loss (unconditional), and MSE + perceptual loss (class-conditioned benign/malignant).}
	\label{fig:diff-quali}
\end{figure}

Figure~\ref{fig:diff-quali} provides a qualitative comparison of the different synthetic image generation strategies. The results illustrate how the inclusion of perceptual loss enhances the visual fidelity and structural representation of skin lesions, while class conditioning further improves the realism and diversity of generated samples.

\begin{table}[]
\label{tab:ham10000}
\caption{Evaluation on HAM10000 for categorical and binary lesion classification. Results are shown for ViT-L and ViT-B baselines and their variants with MAE pretraining (MAE), synthetic data (synth), and knowledge distillation (distill). Comparisons include Derm-t2im~\cite{farooq2024derm} and EfficientNet baselines, as well as EfficientNet models augmented with MAE pretraining and distillation.
}
\resizebox{\columnwidth}{!}{\begin{tabular}{c|cc|cc}
 & \multicolumn{2}{c|}{Categorical} & \multicolumn{2}{c}{Malignant/Benign} \\
Model & ACC & F1 & ACC & F1 \\ \hline
ViT-L (Baseline) & 0.6579 & 0.6383 & 0.7612 & 0.4043 \\
ViT-B (Baseline) & 0.6748 & 0.6498 & 0.7225 & 0.3209 \\ \hline
EfficientNet-B0~\cite{tan2019efficientnet} & 0.7976 & 0.6726 & 0.8050 & 0.7563 \\ 
EfficientNet-B3~\cite{tan2019efficientnet} & 0.7160 & 0.7184 & 0.8157 & 0.7330 \\ 
EfficientNet-B7~\cite{tan2019efficientnet} &  0.7913 & 0.7470 & 0.8152 & 0.7361 \\ 
Derm-t2im~\cite{farooq2024derm} (ViT) & - & - & 0.8502 & - \\ 
Derm-t2im~\cite{farooq2024derm} (MobileNet) & - & - & 0.7358 & - \\ \hline
ViT-B + synth & 0.6719 & 0.6557 & 0.7952 & 0.7506 \\
MAE + ViT-H + synth & \textbf{0.8915} & \textbf{0.8911} & 0.8715 & 0.8715 \\
MAE + ViT-L + synth & 0.8382 & 0.8247 & 0.8512 & 0.8361 \\
MAE ViT-H distill to ViT-B (no synth) & 0.7719 & 0.7557 & 0.8782 & 0.8797 \\
MAE ViT-H distill to ViT-B + (non-cond.) synth & 0.7337 & 0.7216 & 0.8278 & 0.7887 \\
MAE ViT-H distill to ViT-B + synth & 0.8186 & 0.7225 & 0.8278 & 0.7887 \\
MAE ViT-H distill EfficientNet-B0 + synth & 0.8058 & 0.7563 & \textbf{0.9051} & \textbf{0.9010}
\end{tabular}}
\end{table}



\subsection{Performance considerations}

The ViT-H provides strong representations but is impractical for mobile or edge deployment due to its high computational and memory demands ($\approx2.5 GB$ model size and $\approx55-60 GFLOPs$ per inference at $256\times256$ resolution). Although ViT-B reduces the parameter count by approximately $8\times$ ($\approx86 M$ parameters), it remains costly for mobile devices, requiring approximately $17-18 GFLOPs$ per inference. In contrast, knowledge distillation enables transferring the representational power of large transformers to lightweight architectures such as EfficientNet-B0, allowing efficient on-device inference with only $\approx0.6 GFLOPs$ and approximately $5 M$ parameters.

\section{Conclusion}

We addressed class imbalance in skin lesion classification by integrating class-conditioned diffusion-based synthesis with MAE pretraining and knowledge distillation. This strategy enables robust representation learning from limited dermatology datasets while allowing lightweight student models to retain strong discriminative performance for both benign and malignant lesions. Our results demonstrate the effectiveness of synthetic data and teacher-student training for efficient and deployable dermatology classification systems.

\section{Compliance with ethical standards}
\label{sec:ethics}

This research study was conducted retrospectively using human subject data made available in open access~\cite{tschandl2018ham10000}. Ethical approval was not required, as confirmed by the attached license for the open-access data.

\section{Acknowledgments}
\label{sec:acknowledgments}

This work was partially supported by INES.IA (National Institute of Science and Technology for Software Engineering Based on and for Artificial Intelligence) www.ines.org.br, CNPq grant 408817/2024-0. The project was supported by the Ministry of Science, Technology, and Innovation of Brazil, with resources from Law No. 8,248, dated October 23, 1991, under the scope of the PPI-SOFTEX, coordinated by Softex and published under RESIDÊNCIA EM TIC 63 – ROBÓTICA E IA – FASE II, DOU 23076.043130/2025-27. 

\bibliographystyle{IEEEbib}
\bibliography{references}

\end{document}